\documentclass[letterpaper, 10 pt, conference]{ieeeconf}  

\IEEEoverridecommandlockouts                              

\usepackage{graphicx} 
\usepackage{amsmath}
\usepackage{amsfonts}
\usepackage{hyperref}
\usepackage{xcolor}
\usepackage{censor}
\usepackage{amssymb}
\usepackage{booktabs}
\usepackage{comment}
\usepackage{censor}

\hypersetup{
    colorlinks=true,  
    linkcolor=black,  
    citecolor=black,  
    filecolor=black,  
    urlcolor=blue     
}

\title{TACTIC: Understanding Tactile Encoders and Conditioning for Contact-rich Robot Manipulation Policies}
\StopCensoring 
\author{\censor{Seongjin Bien$^{*1}$, 
D\'ebora Oliveira Makowski$^{*1}$, 
Carlo Kneissl$^{*2,5}$, 
Reihaneh Mirjalili$^{1}$, 
\\Pankhuri Vanjani$^{3}$, 
Rudolf Lioutikov$^{3}$, 
Gitta Kutyniok$^{2,5,6}$,  
Florian Walter$^{4}$ 
and Wolfram Burgard$^{1}$}
\thanks{\censor{$^*$Core contributors}.}%
\thanks{\blackout{$^{1}$University of Technology Nuremberg, Germany. }}%
\thanks{\blackout{$^{2}$Ludwig-Maximilians University Munich, Germany.}}
\thanks{\blackout{$^{3}$Karlsruhe Institute of Technology, Germany.}}
\thanks{\blackout{$^{4}$Deggendorf Institute of Technology, Germany.}}
\thanks{\blackout{$^{5}$Munich Center for Machine Learning (MCML), Germany.}}
\thanks{\blackout{$^{6}$University of Tromso, DLR-German Aerospace Center}}
\thanks{Available at \url{https://utn-air.github.io/TACTIC}}
}

\begin{document}
\bstctlcite{IEEEexample:BSTcontrol}

\maketitle

\begin{abstract}
Tactile information is essential for contact-rich manipulation tasks in robotics.
Vision-based tactile sensors make it particularly easy to design end-to-end manipulation policies with tactile sensing, as they enable the use of existing encoders from computer vision.
However, this has led to a huge variety of architectures, training datasets, and evaluation protocols, making it difficult to determine which design choices best encode touch.
In this work, we address this gap and present a comprehensive study of tactile encoders and fusion strategies across various contact-rich manipulation tasks in real-world experiments.
To enable a controlled comparison, we train and evaluate all models under the same pipeline and experimental setup, comprising more than 2000 real-world rollouts.
Our results go beyond other studies that only compare simulation performance, which does not necessarily translate to real-world settings, where large-scale evaluations are needed to obtain reliable statistics.
Our key finding is that there is no universally optimal representation or fusion strategy for encoding visual-tactile. 
Instead, the best encoder backbone and fusion scheme depend strongly on the task.

\end{abstract}

Tactile sensing has become increasingly common in robotic manipulation, as it provides contact information not accessible through third-view cameras.
This has led to a growing variety of tactile encoders, many of which report strong performance on downstream tasks, such as material classification~\cite{agrawal2025t3,anytouch_2026}, pose estimation~\cite{xu2025unit}, and shear-force measurements~\cite{higuera2025sparsh}.
But while most studies show that tactile sensing can deliver strong performance, only a few assess how to integrate it into robotic policies systematically~\cite{luu2026manifeel}.

\begin{figure}[!t]
    \centering
    \includegraphics[width=0.85\linewidth]{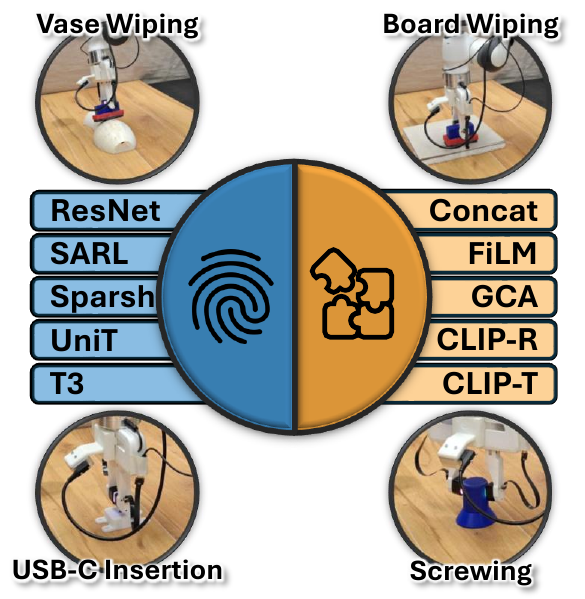}
    \caption{An overview of TACTIC. We study the effect of five tactile encoders paired with five multimodal fusion methods, tested on four different challenging contact-rich manipulation tasks.}
    \label{fig:placeholder}
\end{figure}
Most works that incorporate tactile information into policy training compare task performance solely against vision-only pipelines, not against other tactile encoders~\cite{zorin2026taco}. Even if papers using different tactile encoders report success rates for similar tasks, it is difficult to make a fair comparison, given the different experimental protocols and environments in which these experiments were conducted.
Additionally, most studies perform experiments and evaluations primarily in simulation, making it unclear whether the results translate to the real world~\cite{tacsl}. This is because existing tactile simulations cannot yet capture the complexities and physical properties of the real world, which complicates sim-to-real transfer. On the other hand, works that conduct real-world experiments are typically limited in scale and rarely evaluate robustness to different tasks. As a result, we still lack a clear understanding of which tactile representations are effective for which manipulation tasks.
\vspace{1em}

Despite growing interest in tactile representation learning, the effects of encoder choice and policy conditioning on downstream manipulation performance remain comparatively underexplored. In particular, relatively few studies evaluate multiple tactile encoders under a shared real-world policy and task setting, especially across tasks with different tactile demands. Reproducible comparisons are also made more difficult by the limited availability of open-source real-world benchmarks for this setting.

In this view, we introduce TACTIC, a comprehensive real-world evaluation of multiple tactile encoders under identical training and deployment conditions. To understand when tactile information matters more than vision, we propose a set of challenging, contact-rich manipulation tasks that impose critical yet different requirements for general-purpose tactile representations. We train the Action Chunking Transformer (ACT)~\cite{zhao2023act} policy across five state-of-the-art tactile encoders, five multimodal fusion strategies, and four manipulation tasks. Each condition is rolled out twenty times, resulting in a total of 2,180 evaluations across all conditions, including several visual out-of-distribution studies.

In summary, we make the following contributions. First, we conduct the largest-yet real-world experimental evaluation of five state-of-the-art tactile encoders and different fusion methods across four challenging contact-rich manipulation tasks, and test them under controlled experimental setups and protocols. Second, we perform an in-depth analysis of their individual performances, identify possible patterns in the fusion methods, and provide insight into what matters the most. Finally, we make all task materials, training pipelines, and evaluation configurations open-source to support reproducibility and future comparisons.

\section{Related Works}

\subsection{Tactile Sensing for Robot Manipulation}
Prior work has shown that tactile feedback improves performance in contact-rich and visually ambiguous manipulation across several types of single-task imitation policies~\cite{tactilealoha2025,zhao2026vitactracingvisualtactileimitationlearning,heng2026vitacformer,vital2025,gelfusion2025}. The differing temporal characteristics of tactile and visual observations have also motivated slow-fast and hierarchical policies that couple global action generation from higher-rate feedback~\cite{xue2025reactive,chen2026implicitrdp,foar2025,zheng2026omnivtavisuotactileworldmodeling}. Another line of work has extended tactile sensing into Vision-Language-Action models, exploring heterogeneous tactile modalities, representation alignment, and contact-aware reasoning~\cite{cheng2025omnivtla,zhang2026tacvla,huang2025tactilevla,huang2026tafvla}.

\subsection{Encoding Tactile Information}
In general, vision-based tactile learning has increasingly shifted from generic encoders such as ResNet-18 towards task-tailored architectures. T$^3$~\cite{agrawal2025t3} and AnyTouch~\cite{anytouch_2026} learn sensor-general transformer representations from heterogeneous tactile sensors, while AnyTouch 2~\cite{anytouch2_2026} extends this direction toward dynamic tactile perception by emphasizing temporal contact evolution. Sparsh~\cite{higuera2025sparsh} uses large-scale self-supervised ViT pretraining on sensors including DIGIT~\cite{digit} and GelSight~\cite{gelsight}, with temporal objectives designed to capture contact dynamics. UniT~\cite{xu2025unit} instead learns compact tactile latents through VQGAN-based pretraining, whereas SARL~\cite{khurana2025sarl} combines a ResNet-18 backbone with spatially aware self-supervision. We evaluate a representative subset of these approaches; the selection criteria are detailed in Section~\ref{sec:method}.

Within single-task imitation policies, tactile information has been integrated through a wide range of fusion strategies, including direct concatenation~\cite{tactilealoha2025}, FiLM conditioning~\cite{huang2026sharpa}, cross-attention~\cite{gelfusion2025,heng2026vitacformer}, learned gating~\cite{ruan2026retacact}, and contrastive visuo-tactile representation learning~\cite{vital2025,freetacman2026,liu2026crossmodalvisuotactilerepresentationlearning,convitac2025}.

\subsection{Benchmarking Tactile Manipulation} Benchmarks such as ManiFeel~\cite{luu2026manifeel}, TacO~\cite{zorin2026taco}, and CONTACT~\cite{saka2026contactcontactawaretactilelearning} study the effect of tactile sensing and representation choices across manipulation tasks, while tactile encoder works such as AnyTouch2~\cite{anytouch2_2026} include downstream policy ablations. However, the interaction between the tactile backbone and the fusion strategy remains comparatively underexplored within a common policy and evaluation environment. 

We address this gap through a controlled ablation over tactile encoders and fusion mechanisms using ACT~\cite{zhao2023act}. ACT is widely used in tactile imitation learning, and prior work suggests that tactile-driven gains can follow similar trends under Diffusion Policy~\cite{vital2025}. We exclude VLAs and hierarchical systems from the main ablation because they introduce additional design dimensions, such as modality-specific tokenization, representation alignment, temporal scheduling, and auxiliary control pathways, which make it difficult to isolate the effects of tactile representation and fusion.

\section{Method\label{sec:method}}

\subsection{Data Collection and Inference Setup}
Our setup consists of two Franka FR3 robot arms, which enables bilateral leader-follower style teleoperation with haptic feedback capabilities. 
The leader robot is controlled by the teleoperator in gravity compensation mode, and the follower robot tracks the leader robot's joint positions while its external torque readings accessible through \texttt{libfranka} are added to the leader robot's joint torques:
\begin{align}
    \tau_{fol} &= K_p(q_\text{lead} - q_\text{fol}) - K_d \dot{q}_\text{fol} \\
    \tau_{lead} &= -G_h \cdot \tau_\text{ext, fol}
\end{align}
where $K_p$ and $K_d$ are the joint-level stiffness and damping parameters, $q_\text{fol}$ is the joint state vector of the follower (i.e., proprioception), $q_\text{lead}$ is the joint state vector of the leader (i.e., action), and $G_h$ are the haptic feedback gains, adjusted between $[1, 1.5]$ depending on the task. 

The follower robot is equipped with a Franka Hand with custom fingers to rigidly attach two DIGIT sensors and a Realsense D405 wrist camera. A Realsense D435 camera is positioned diagonally to provide a full view of the follower robot and its workspace. During inference, the follower robot is used along with all modalities. We denote the sampled action $
\mathcal{A}$ from the policy $\pi_\theta$ as
\begin{equation}
    \mathcal{A}_t\sim\pi_\theta(I_{\text{DIGIT}_R}, I_{\text{DIGIT}_L}, I_{\text{wrist}}, I_{\text{side}},q)
\end{equation}
where $I$ is an image of the respective sensors and $q$ is the proprioceptive joint and gripper state.

\subsection{Tasks}

\begin{figure}
    \vspace{5.1pt}
    \centering
    \includegraphics[width=0.98\linewidth]{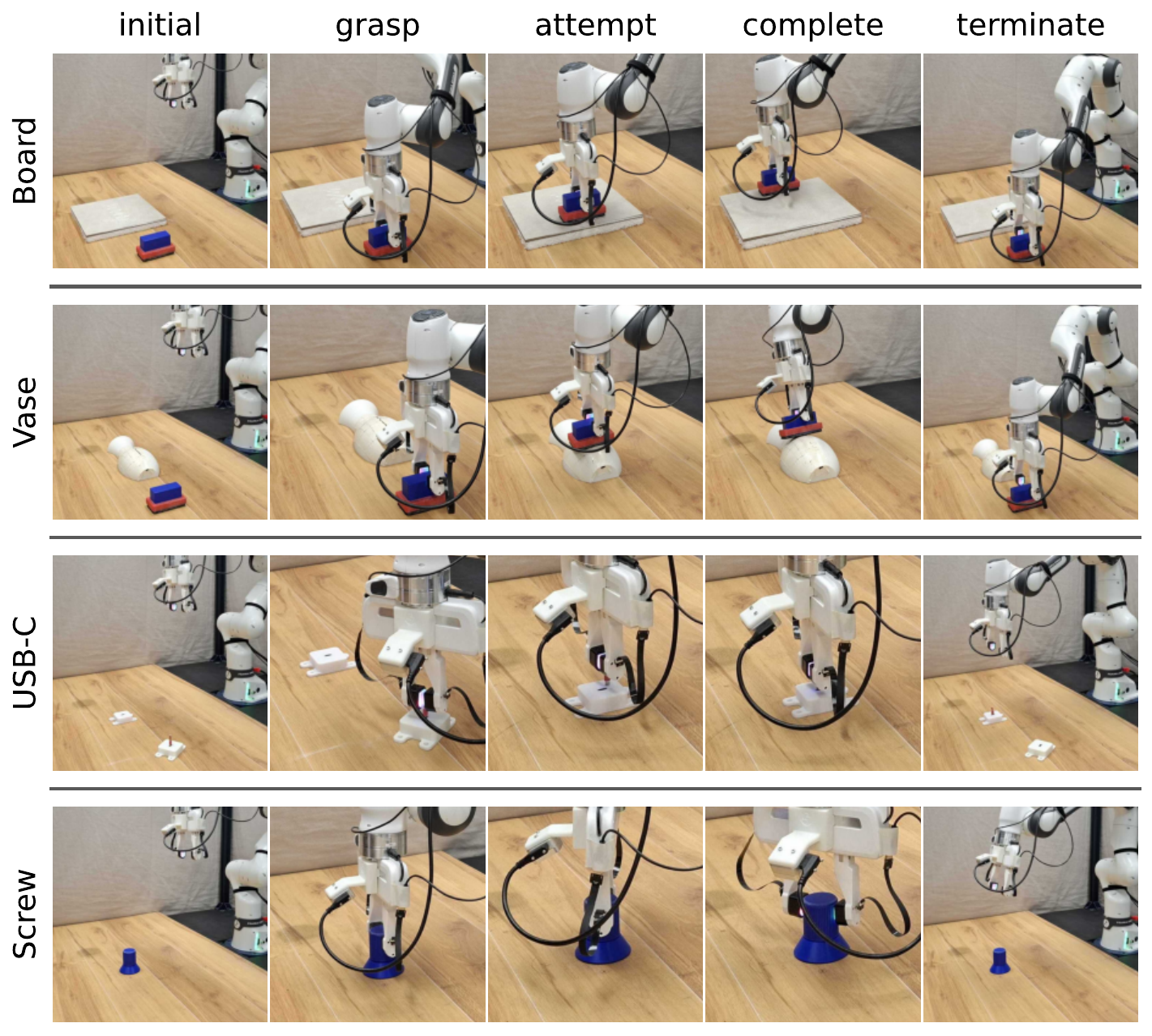}
    \caption{The 4 tasks tested in TACTIC. Each task has 5 stages, allowing for a more fine-grained analysis of the failure points for each individual condition.
    The success criterion for the wiping tasks is to leave no residue on the board or vase surface. 
    For USB-C insertion, success is counted when the USB is inserted in the correct position, but not necessarily forced in.
    For screwing, success is defined as fully tightening the nut.}
    \label{fig:tasks}
\end{figure}
To assess different aspects of vision-based tactile perception, we design four contact-rich manipulation tasks, as depicted in~\autoref{fig:tasks}. 
Success rates are tracked across five different stages: initialization, grasp, task attempt, task completion, and task termination. 
For each task, we collect approximately 100 demonstrations at 30~Hz, using four different pairs of DIGIT cartridges to increase the diversity of tactile appearance and reduce overfitting to cartridge-specific characteristics.

\subsubsection{Board wiping}
The robot grasps an eraser, wipes a chalk line from a board, and places the eraser back down. This task primarily tests sensitivity to normal contact forces. 
For each demonstration, we vary the chalk line's pose and orientation, as well as the board height.

\subsubsection{Screwing}
The robot tightens a pre-positioned nut and then retreats to the initial pose. Because there is no clear visual signal for task completion, the detection of the termination signal must rely on tactile feedback to identify the shear/slip characteristics associated with a fully tightened nut. The required number of turns varies across demonstrations.

\subsubsection{USB-C insertion}
The robot picks up a USB-C plug, inserts it into a fixed port, and then returns to its initial pose. This task examines whether tactile information improves high-precision, low-tolerance insertion and recovery from small alignment errors. The environment remains fixed. Variation arises from small differences in the plug's pose within the gripper that are introduced during teleoperation.

\subsubsection{Vase wiping}
The robot wipes a marked line from the curved surface of a 3D-printed vase bisected along its length, and returns the eraser. 
In addition to controlling the normal force, this task requires reasoning about the relationship between grasp pose and the eraser's contact geometry on a curved surface. The line position is varied around the vase's circumference.

\subsection{Encoder Backbone}

We evaluate five tactile backbones, with weights shared across the left and right tactile sensors. The wrist and side-camera streams are always encoded with a shared ResNet-18~\cite{he2016deep} initialized from ImageNet~\cite{russakovsky2015imagenet} and fine-tuned end-to-end. Unless otherwise noted, we estimate a background image from 50 blank tactile frames at the beginning of each rollout and subtract it from the tactile stream to reduce sensitivity to cartridge-specific appearance.

As a baseline for the tactile backbone, we use an ImageNet-initialized ResNet-18 and fine-tune it jointly with the action expert. We then evaluate four tactile-specific encoders spanning distinct pretraining strategies. \textbf{SARL}~\cite{khurana2025sarl} uses a ResNet-based self-supervised teacher-student formulation with spatially aware objectives; we pretrain SARL from scratch on our tactile demonstrations. \textbf{UniT}~\cite{xu2025unit} learns a quantized tactile latent space using a VQ-GAN objective~\cite{esser2021taming}; we initialize it from the released pretrained weights and fine-tune it on our data. \textbf{Sparsh}-DINO~\cite{higuera2025sparsh,caron2021emerging} uses a ViT-based self-supervised encoder; following the released DIGIT configuration, we concatenate frames at timesteps $t$ and $t-4$ along the channel dimension to provide short-horizon temporal context, and fine-tune it on our data. \textbf{T$^3$} uses sensor-specific encoders with a shared latent trunk across heterogeneous tactile sensors. We use its medium DIGIT variant with the released pretrained weights, and use it as-is. Unlike the other encoders, T$^3$ is applied to raw DIGIT images without background subtraction to match its pretraining configuration. 

These backbones were selected to cover complementary design choices: end-to-end CNN training (ResNet-18), tactile-specific CNN self-supervision (SARL), discrete generative pretraining (UniT), large-scale self-supervised ViT pretraining with temporal context (Sparsh-DINO), and heterogeneous multi-sensor pretraining (T$^3$).

For the pretraining or adaptation of the tactile encoder, we use only tactile images from the collected demonstrations. To reduce the dominance of no-contact observations, 95\% of the sampled training frames are drawn from periods in which the gripper is closed and 5\% from periods in which it is open. During action expert training, the pretrained tactile encoders are frozen. Only the baseline ResNet tactile encoder is optimized end-to-end with the policy.

\subsection{Multimodal Feature Fusion}
We evaluate five multimodal fusion strategies. \textbf{Concat} directly concatenates visual and tactile features along the channel dimension. When needed, a 1D convolution projects the tactile representation to the visual feature dimension. \textbf{FiLM} conditions the tactile representation on the binary gripper state. A two-layer MLP with GELU activation predicts per-channel scale and shift parameters, which modulate the tactile features before fusion. This provides the policy with an explicit contact prior, since meaningful DIGIT deformation in our tasks primarily occurs while the gripper is closed. 
 
For \textbf{gated cross-attention} (GCA), tactile tokens query visual tokens through a single multi-head cross-attention layer, with visual tokens serving as keys and values. The resulting update is added to the original tactile representation through a learned gate. Unlike FiLM, this fusion mechanism uses no proprioceptive conditioning.  

The remaining two variants use CLIP-style contrastive pretraining and differ only in how positive and negative samples are constructed. The vision encoder remains frozen. For the ResNet-18 tactile baseline, the tactile encoder is optimized during contrastive pretraining. For the pretrained tactile backbones, the backbone remains frozen, and only a $1\times 1$ convolution projection layer is learned. Visual and tactile features are mapped through modality-specific projectors, both again two-layer MLPs with GELU activation, and optimized using a symmetric CLIP objective. Following prior works~\cite{vital2025,freetacman2026, liu2026crossmodalvisuotactilerepresentationlearning}, the projectors are discarded after pretraining, and the resulting tactile encoder and projection layer are frozen during ACT training.

For \textbf{CLIP-R}, each batch contains $N=256$ samples, divided equally across the $D=4$ task datasets, with samples drawn randomly from distinct trajectories. For \textbf{CLIP-T}, we follow the time-aware sampling of VITaL~\cite{vital2025}: batches contain $N=32$ samples, again balanced across tasks, with samples from each task drawn from the same trajectory at a minimum stride of $s=30$ frames (1 second). In both cases, half of the samples are drawn from closed-gripper states and half are uniformly drawn from the full trajectories. Comparing CLIP-R and CLIP-T isolates whether temporally structured contrastive sampling provides additional benefit over random cross-modal alignment.
 
\subsection{Action Expert}
Action Chunking with Transformers (ACT)~\cite{zhao2023act} is a widely used policy for training end-to-end robot policies with tactile sensing.
It is a transformer-based architecture that takes in all camera-feed observations and sensor data to predict a sequence of future actions, i.e., a chunk.
An action step comprises the delta joint positions to execute for the next $T$ timestamps.
In this paper, we trained all policies with a 120-future-step horizon.
Unlike VLAs, each ACT instance is specialized to a single dataset. 

To improve smoothness, ACT applies a weighted average of past predicted actions after querying a new trajectory. 
This temporal ensembling merges future action points predicted for the same timestep but not yet executed.
Note that the exponential-average weight directly affects the precision of the executed task. 
Stronger averaging reduces small differences between consecutive action chunks and is therefore useful for precise tasks. If the averaging is too weak, the motion can become jerky and less accurate. We used temporal ensembling coefficient of 0.01 for all tasks except USB-C, where 0.1 was used to prevent the policy from getting stuck.

\section{Experiments}

To fairly compare the different combinations of backbones and multimodal fusion schemata, we ran 20 real-world inference rollouts per policy ablation condition.
For 5 backbones, 5 fusion schemata, and 4 tasks, including out-of-domain experiments, we conducted 2,180 rollouts in total.
To evaluate the impact of tactile data on each task, we also ran a vision-only policy conditioned on the side and wrist camera encoders, i.e., without tactile input. 
Our experiments focus on two key questions:
\begin{enumerate}
    \item Which tactile encoders generalize across tasks with different tactile requirements, which are task-specific?
    \item How do different multimodal fusion strategies affect the downstream policy?
\end{enumerate}

\begin{figure*}
    \vspace{5.1pt}
    \centering
    \includegraphics[width=\linewidth]{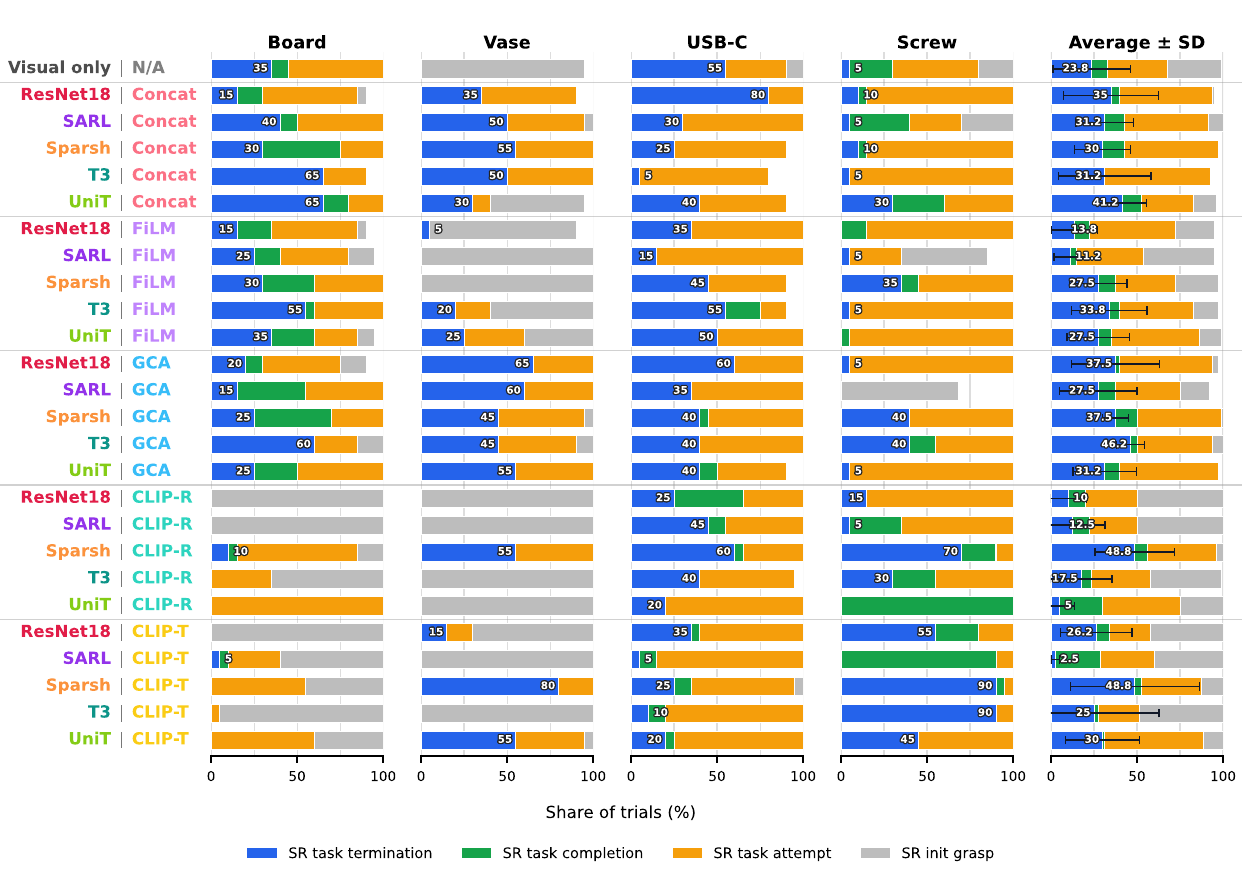}
    \caption{Success rates of initial grasping, task attempt, task completion, and task termination for the five ablated backbones and fusion schemata for the ACT policy and each proposed dataset. The first line shows the closed-loop metrics for the vision-only (no tactile) case. The numbers on the plot represent the task-termination success rate. If no number is shown, the success rate of task termination is zero. The average column also includes the standard deviation of the termination success rate across the different tasks.}
    \label{fig:big-fig-per-backbone}
\end{figure*}

\subsection{In-domain Experiments}
We report the success rate per initial grasp, task attempt, completion, and termination for each ablation in Fig.~\ref{fig:big-fig-per-backbone}. On average, Sparsh-CLIP variants have the highest successful termination rate at almost 50\%, but are slightly outperformed by UniT-Concat in task completion. Additionally, both Sparsh-CLIP variants perform extremely poorly on the board task and are surpassed on the USB-C task by the simplest tactile baseline, ResNet18-Concat.
These results indicate that strong average performance can conceal substantial task-level variation: when paired with the right fusion strategies, tactile encoders behave as strong task specialists. This necessitates task-level analysis to reveal how these specialization patterns arise from the individual encoder and fusion design choices.

\subsubsection{Board Wiping}
On the board task, the CLIP variants perform significantly worse than the alternatives, often failing to progress much beyond the grasp stage. Board wiping is primarily a planar, repetitive contact-maintenance task in which task completion depends heavily on the wrist camera, since the white chalk is barely visible from the side view. Tactile feedback is mainly useful for maintaining contact and sufficient pressure against the board. During rollout, we observed that many CLIP-R variants released the eraser after lifting it up, with only a small fraction progressing to wiping. Even when they did, their wiping motions exhibited a limited range, with the policy tending to execute similar trajectories over the same region of the board, regardless of where or how the lines were drawn. 

In contrast, the non-CLIP variants performed well, adapting their wiping trajectories accordingly. 
T$^3$ performs well with all non-CLIP fusion strategies, while SARL, Sparsh and UniT work best with Concat. 
Except for ResNet18, Concat shows the highest completion rate.
We hypothesize that this difference in behavior between CLIP and non-CLIP strategies is caused by the large variation in wiping locations. 
Aligning relatively similar tactile representations to a diverse distribution of visual embeddings through a shared projection may be restrictive. It is thus difficult to learn a cross-modal representation that remains sensitive to spatial variation in the visual input.

\subsubsection{Vase Wiping}
On the vase task, many of the FiLM and CLIP variants, as well as the vision-only policy, did not progress beyond the initial grasp, dropping the eraser immediately after lifting it. This behavior resembled the failure mode observed for the CLIP variants of the board task. The vase task requires the policy to place the eraser back down at its starting position, which occludes the vase from the side camera view. We suspect this makes the representations between the grasping and termination stages hard to distinguish. Given that ACT does not have any form of closed-loop feedback or task memory that allows it to track previous states or actions, this similarity may cause the model to mix grasping and releasing.

Nevertheless, Concat and GCA variants overcome these ambiguous representations. 
This suggests that the underlying encoder representations already contain useful task-relevant information without requiring a more complex fusion mechanism. 
GCA can further selectively modulate tactile information, thereby increasing the relevance of tactile representations during contact phases of the task.

Sparsh is a major outlier in this task. It seems to benefit greatly from the CLIP-T strategy and shows limited degradation when paired with CLIP-R. This suggests that the temporal nature of this encoder provides a sufficiently distinct signal between the beginning and end of the task. UniT also stands out. UniT-CLIP-R fails to progress beyond the initial grasp, similar to other CLIP-R variants, whereas UniT-CLIP-T achieves the same 55\% termination rate as UniT-GCA. It is also more stable than the UniT-Concat variant, which frequently shows the dropping failure mode. These results reinforce that no single fusion strategy is consistently optimal across encoders.

\subsubsection{USB-C Insertion}
The differences between fusion strategies are less clear on this task. ResNet18-Concat, despite being the simplest tactile variant, performs best, achieving 80\% task termination rate. More complex fusion strategies consistently reduce ResNet18's performance. However, for the other encoders, FiLM, GCA, and CLIP-R have more heterogeneous effects. T$^3$ benefits substantially from FiLM conditioning, reaching a 55\% termination rate and a 75\% completion rate. A similar improvement under FiLM is observed for Sparsh and UniT, whereas SARL experiences a drop in performance.

During rollout, we observed that both CLIP-R and CLIP-T variants produced jerky and sporadic motions. CLIP-R often had enough accuracy to bring the plug close to the port, whereas CLIP-T overshot the port and was unable to recover. 
Although VITaL~\cite{vital2025} reported a 20\% improvement using CLIP-T over the vision-only baseline in their setup with four cameras, we could not reproduce this result. We hypothesize that our 2-camera setup cannot fully exploit the CLIP-T mechanism. 

Given that the vision-only variant already achieves a 55\% termination rate on this task, these results suggest that tactile information is not uniformly beneficial for USB-C insertion. Instead, poorly integrated tactile features may interfere with visual information, making architecture choices especially important for this task.

\subsubsection{Screwing}

\begin{figure}[htbp]
    \vspace{5.1pt}
    \centering
    \def\svgwidth{.92\columnwidth}
    \includegraphics[width=\linewidth]{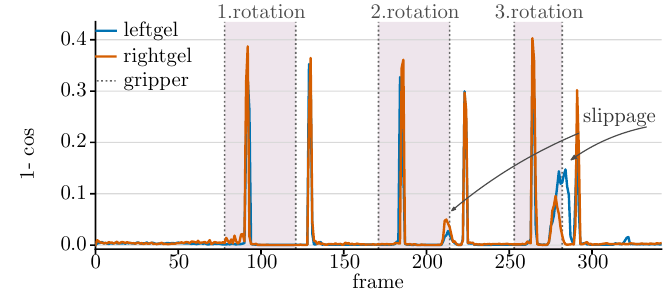}
    \caption{Cosine similarity between the Sparsh embeddings obtained in the regular way, i.e., by feeding tactiles at time $t$ and $t-4$ into the encoder and those without temporal information, i.e. feeding both times the $t$ tactile into the model.}
    \label{fig:screw-sparsh-temporal}
\end{figure}

\begin{table}[t]
  \centering
\caption{Linear probing \cite{alain2017probes} of tactile encoders to discriminate between the last and second-to-last turn on the screw task.  }
  \small
  \label{tab:last-turn-detector}
  \begin{tabular}{c|cccc}
    Method & Sparsh & T$^3$ & SARL & UniT \\
    \midrule
    $R^2$ & 0.138 & 0.174 & 0.055 & 0.036  
   \end{tabular}
\end{table}
On the screw task, CLIP-T variants perform the best by far, with nearly every encoder benefiting from this fusion strategy, except SARL. CLIP-T encourages temporal alignment of the features, which is especially important for capturing slippage/shear force signals once the cap is tight.
Both Sparsh-CLIP variants perform well on this task, with Sparsh-CLIP-T achieving a best-in-task termination rate of 90\%. This result is consistent with Sparsh's architecture, as it is the only encoder that explicitly incorporates temporally separated tactile observations. This makes Sparsh particularly suited to recognizing changes in shear forces in the tactile images, as shown in Fig.~\ref{fig:screw-sparsh-temporal}. 

T$^3$-CLIP-T also stands out, achieving the best termination rate of 90\% despite not being fine-tuned on our dataset, unlike the other encoders. One possible explanation lies in T$^3$'s pretraining objectives, which include material texture classification and pose estimation. Both tasks require sensitivity to subtle variations in tactile observations, which may help preserve the shear-force and object-slip cues that emerge once the cap is fully tightened. A probing analysis \cite{alain2017probes} confirms that T$^3$ and Sparsh can better detect when the nut is tightened, as shown in Table~\ref{tab:last-turn-detector}.

\subsubsection{Effect of CLIP Fusion}

With the task-level overview at hand, we now perform in-depth analysis of the more notable behaviors during rollout, particularly the impact of the CLIP-based methods.
They seem to have the most substantial impact on performance among all fusion strategies considered. An important point to mention is that the only difference between the raw tactile embeddings and their CLIP counterparts is a linear transformation, except for ResNet18, which is optimized during contrastive pretraining.

\begin{table}[t]
\vspace{5.1pt}
  \centering
  \caption{Effect of the contrastive adapter on $\frac{\operatorname{PR} \left(\Sigma_{\text{CLIP-\{R,T\}}} \right)}
{\operatorname{PR}(\Sigma_{\text{pre}})}$ of the pooled tactile
  representation.}
  \small
  \label{tab:participation}
  \begin{tabular}{@{}lcc@{}}
    \toprule
    Encoder & Screw (CLIP-R / CLIP-T) & Vase (CLIP-R / CLIP-T) \\
    \midrule
    SPARSH & \textbf{5.16} / \textbf{4.25} & \textbf{3.40} / \textbf{1.60} \\
    T$^3$     & 1.10 / 0.65                   & 0.80 / 0.50 \\
    SARL   & 1.22 / 1.05                   & 0.71 / 0.40 \\
    UniT   & 1.02 / 1.30                   & 1.03 / 1.27 \\
    \bottomrule
  \end{tabular}
\end{table}
To assess the differences between using CLIP-trained and raw embeddings in policy training, we analyze them in the context of vase and screw tasks, as they represent the contrastive directions in which CLIP strategies influence task performance relative to other fusion methods. 
On screw, CLIP methods increase performance from 40\% up to 90\%, whereas on vase most encoders collapse, except Sparsh. 

It is thus insightful to view the effect of the 1-D projection layer trained with CLIP from the perspective of a Principal Component Analysis (PCA). 
We hypothesize that distributing embedding variance across more principal components helps to reveal more diverse features for the downstream policy to consume.
We measure this using the participation ratio \cite{litwinkumar2017optimaldegrees} defined as
\begin{equation}
    \operatorname{PR}(\Sigma)=\frac{\bigl(\sum_i\lambda_i\bigr)^2}{\sum_i\lambda_i^2}
\end{equation}
for a covariance matrix $\Sigma$ and its non-increasing eigenvalues $\lambda_i$, which can be interpreted to measure the number of active components.
Specifically, we consider the ratio
$\frac{\operatorname{PR}(\Sigma_{\text{CLIP}})}
{\operatorname{PR}(\Sigma_{\text{pre}})}$,
where $\Sigma_{\text{CLIP}}$ denotes the empirical covariance of the tactile embeddings after the CLIP transformation and $\Sigma_{\text{pre}}$ the one before.
Indeed, the results in Table~\ref{tab:participation} back up this claim.
It demonstrates a decreased participation ratio for for SARL and T$^3$, indicating that more variance collapses onto fewer principal components, whereas for Sparsh and UniT the PR actually increases.

For the screw task, we see an increase in PR of over $1.0$ for both Sparsh-CLIP-T and Sparsh-CLIP-R. 
Initially, T$^3$-CLIP-T appears to be a counterexample to this hypothesis, as its PR decreases while its performance increases.
However, actually correlating the first principal component of the T$^3$ embedding with the tactile contact magnitude, the correlation jumps from $0.16$ to $0.75$ after the CLIP transformation.
Thus, CLIP helps to re-rank the already existing signal in the T$^3$ embeddings, which can help against the primary failure mode on the screw task, which is premature stopping.

In summary, CLIP only transforms the underlying signal but cannot add information. 
The PCA analysis helps to understand how CLIP modifies the tactile signal and whether the signal expanded or collapsed.
Nevertheless, it ultimately cannot make predictions about downstream performance, which depends crucially on the task and tactile signal. 

\subsection{Out-of-Domain Analysis}

\begin{figure}[t]
    \vspace{5.1pt}
    \centering
    \includegraphics[width=0.7\linewidth]{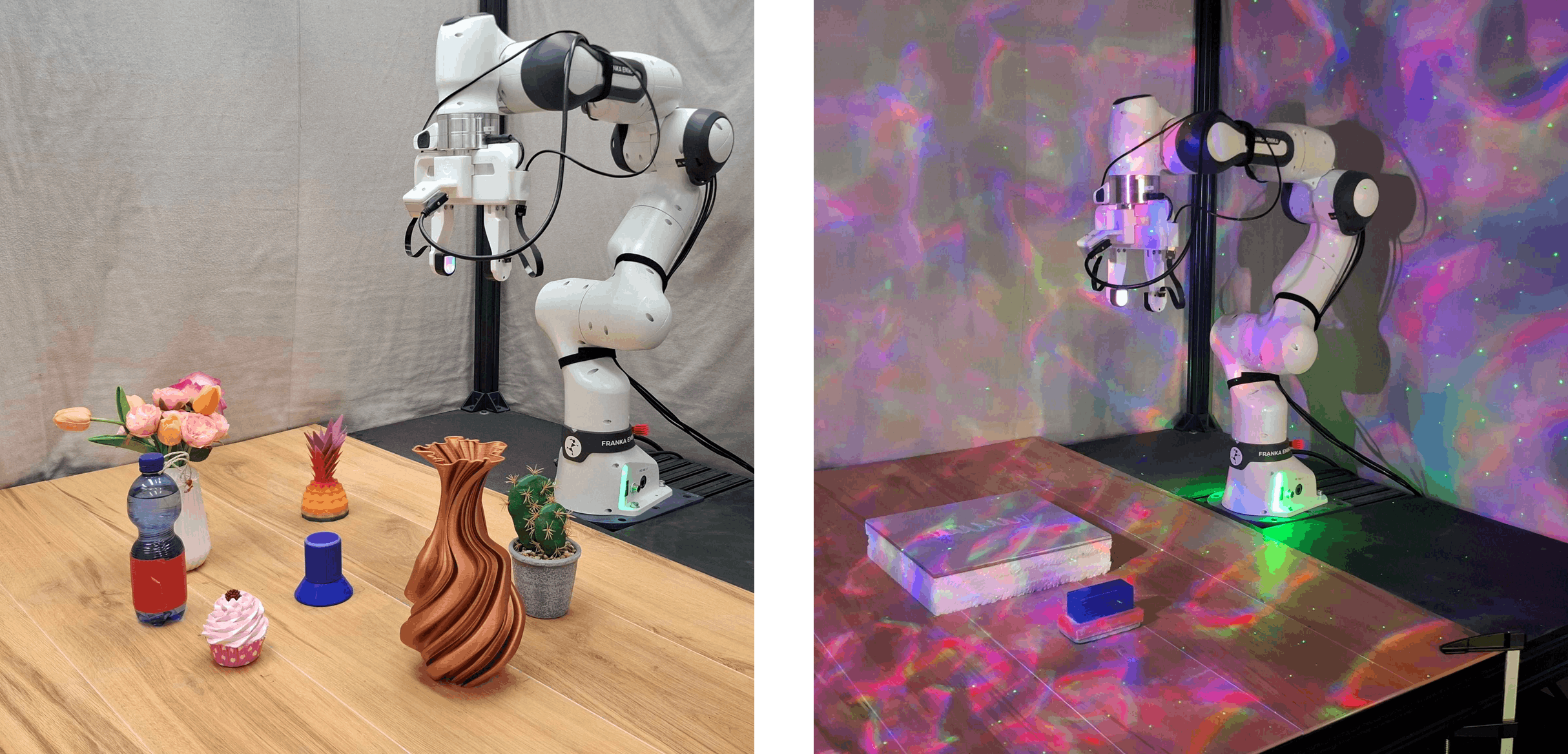}
    \caption{The two visual out-of-distribution conditions best models were tested under. Left: Setup with distractor objects. Right: Setup with darkened lights and moving disco lights.  }
    \label{fig:visual_ood}
\end{figure}

\begin{figure}[t]
    \centering
    \includegraphics[width=\linewidth]{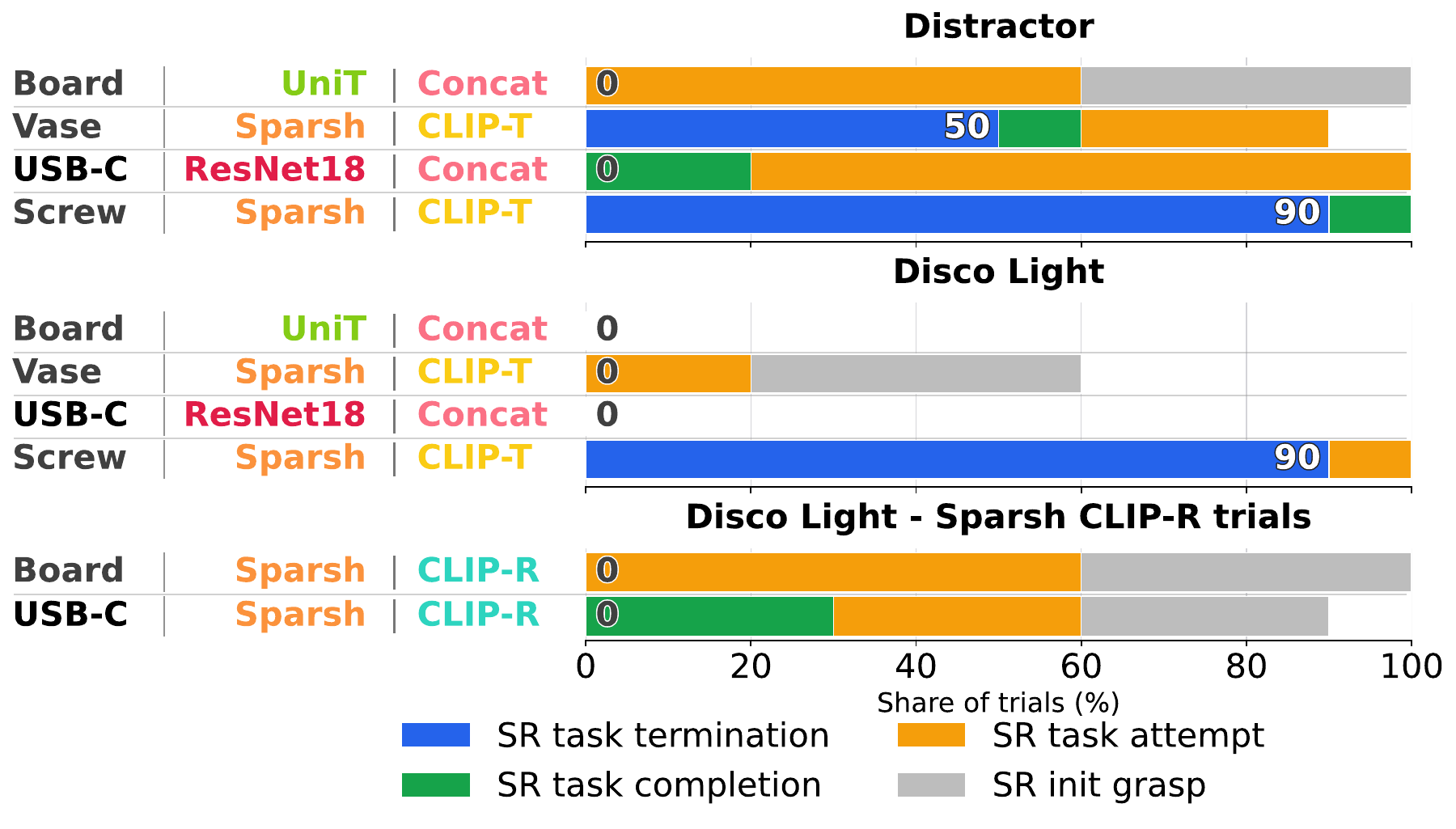}
    \caption{Stage-level success rates of best-in-task model variants under out-of-distribution conditions. Top: Distractor condition. Middle: Disco light condition. Bottom: Best-in-task CLIP variants under disco light condition.}
    \label{fig:disco_ood_evaluation}
\end{figure}

\begin{figure}[t]
    \vspace{5.1pt}
    \centering
    \includegraphics[width=0.9\linewidth]{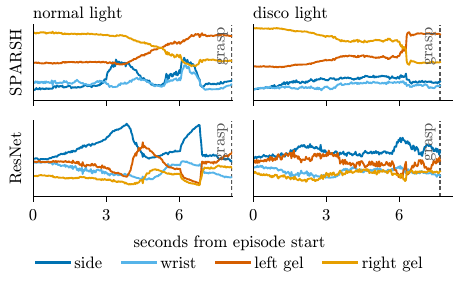}
    \caption{Attention score plots of the approach phase of the USB-C under normal lighting and disco lighting conditions for the Sparsh-CLIP-R (top row) and ResNet-Concat (bottom row) configuration.}
    \label{fig:usbc_disco_pregrasp_attn_comparison}
\end{figure}

To test out-of-distribution (OOD) behavior, we artificially create a visual distribution shift in our experiments by introducing disco-light and distractor objects, as shown in Fig.~\ref{fig:visual_ood}.
For each OOD condition, we evaluate the best-performing policy variants for each task 10 times. The results are shown in Fig.~\ref{fig:disco_ood_evaluation}.

Under the distractor condition, the non-CLIP variants were able to successfully grasp the target objects, but failed to achieve a successful termination. On the board task, UniT-Concat could not localize the chalk line. On the USB-C task, ResNet18-Concat attempted the insertion and even succeeded in two cases, but its performance degraded significantly. On the other hand, Sparsh-CLIP-T performed well. It managed to achieve 50\% termination rate on the vase task, and matched the in-distribution performance on the screw task. 

The disco-light condition showed a similar yet more pronounced pattern. Both non-CLIP variants failed completely in their respective tasks, being unable to even grasp their target objects, while Sparsh-CLIP-T fared better. On the screw task, it matched in-distribution performance despite the highly adversarial condition.
We hypothesize that this is due to the non-CLIP variants being highly dependent on the vision encoder, which cannot produce meaningful embeddings under OOD settings. Sparsh-CLIP-T attends less to vision embeddings, thus placing more importance on proprioception and tactile embeddings. This allows it to still generate accurate trajectories.

To test this hypothesis, we conducted two additional experiments using the best-performing CLIP variants for the board and USB-C tasks under the disco-light condition. Fig.~\ref{fig:disco_ood_evaluation}, bottom shows the rollout results. Despite not having any successful termination, Sparsh-CLIP-R nonetheless manages to grasp the eraser and attempts the board task in more than half of the cases, and on the USB-C task, manages to insert it 3 out of 10 times, supporting our hypothesis.

We also recorded the attention masses of the ACT action decoder during the approach phase of the USB-C task from held-out demonstrations under both nominal and disco-light conditions. 
Fig.~\ref{fig:usbc_disco_pregrasp_attn_comparison} shows these attention curves in the first phase of the USB-C task, i.e., until the first grasp occurs.
Under normal conditions, the ResNet18-Concat variant's decoder primarily attends to the scene camera in order to navigate to the precise location from which to pick up the USB-C cable. This signal is destroyed in the disco-light setting, which explains the policy's failure. On the other hand, Sparsh-CLIP-R attends less to the cameras overall, which means it is therefore not affected as much by the disco light's distribution shift, allowing it to produce its regular approaching trajectory with a higher probability.

\section{Conclusion}
We presented TACTIC, a systematic real-world study of five tactile encoders and five multimodal fusion strategies for ACT policies, comprising 2,180 rollouts across four contact-rich manipulation tasks.
We show that no encoder–fusion combination is best across all tasks, and that the right choice depends on the tactile signal the task requires.
Sparsh and T³ variants performed the best on average, showing that 
tactile representations benefit from large-scale pretraining involving different sensors. 
We also found evidence that temporal information produces richer signals in tasks requiring shear-forces and dynamic tracking.
Regarding fusion strategies, we surprisingly show that simple concatenation is still a competitive approach in relation to more complex conditioning mechanisms.
We demonstrate that contrastive-based strategies induce specialist behavior in policies rather than generalizing performance across all tasks. 
With PCA analysis, we show that this fusion redistributes variance, or promotes a cue that was already present, but it cannot add information.
Finally, our out-of-distribution experiments using attention maps provide empirical evidence that contrastive alignment lets policies attend more evenly to vision and tactile.
We hope that this study guides future work toward developing a conditioning mechanism that generalizes performance across different tactile requirements in contact-rich tasks.

\section*{Acknowledgments}
 \blackout{The authors acknowledge the scientific support and HPC resources provided by the Erlangen National High Performance Computing Center (NHR@FAU) of the Friedrich-Alexander-Universität Erlangen-Nürnberg (FAU) under BayernKI provided by Bavarian state authorities. This work has been partially supported by the German Federal Ministry of Research, Technology and Space (BMFTR) under the Robotics Institute Germany (RIG) and GeniusRobot (BMBF grant no. 01IS24083). This paper is supported by the DAAD program Konrad Zuse Schools of Excellence in Artificial Intelligence, sponsored by the Federal Ministry of Research, Technology and Space. G. K. acknowledges support by the German Research Foundation under Grants DFG-SPP-2298, KU 1446/31-1 and KU 1446/32-1, as well as the project ``Next Generation AI Computing (gAIn),'' funded by the Bavarian Ministry of Science and the Arts and the Saxon Ministry for Science, Culture, and the Hightech Agenda Bavaria. } The authors acknowledge the use of ChatGPT and Claude Code for portions of code creation. All generated code was rigorously reviewed, adapted and validated by the authors. All methodologies, evaluations and interpretation of results were designed and performed by the authors.

\bibliographystyle{IEEEtran}
\bibliography{references} 

@IEEEtranBSTCTL{IEEEexample:BSTcontrol,
CTLuse_forced_etal = "yes",
CTLmax_names_forced_etal = "6",
CTLnames_show_etal = "3",
CTLdash_repeated_names = "no"
}

@string{ieeetro = "IEEE Transactions on Robotics"}

@string{ieeeral = "IEEE Robotics and Automation Letters"}

@string{IROS = "Proc.~of the IEEE/RSJ Int.~Conf.~on Intelligent Robots and Systems (IROS)"}

@string{ICRA = "Proc.~of the IEEE Int.~Conf.~on Robotics \& Automation (ICRA)"}

@string{ICCV = "Proc.~of Int.~Conf.~on Computer Vision (ICCV)"}

@string{IJCV = "International Journal of Computer Vision"}

@string{ICML = "Proc.~of the Int.~Conf.~on Machine Learning (ICML)"}

@string{CVPR = "Proc.~of the IEEE Computer Society Conference on
                  Computer Vision and Pattern Recognition (CVPR)"}

@string{rss = "Proc.~of Robotics: Science and Systems (RSS)"}

@string{ICLR = "Proc.~of the Int.~Conf.~on Learning Representations (ICLR)"}

@string{CORL = "Proc.~of the Conf.~on Robot Learning (CoRL)"}

@inproceedings{caron2021emerging,
  title={Emerging properties in self-supervised vision transformers},
  author={Caron, Mathilde and Touvron, Hugo and Misra, Ishan and J{\'e}gou, Herv{\'e} and Mairal, Julien and Bojanowski, Piotr and Joulin, Armand},
  booktitle=ICCV,
  year={2021},
}

@inproceedings{esser2021taming,
  title={Taming transformers for high-resolution image synthesis},
  author={Esser, Patrick and Rombach, Robin and Ommer, Bjorn},
  booktitle=CVPR,
  year={2021}
}

@article{russakovsky2015imagenet,
  title={Imagenet large scale visual recognition challenge},
  author={Russakovsky, Olga and Deng, Jia and Su, Hao and Krause, Jonathan and Satheesh, Sanjeev and Ma, Sean and Huang, Zhiheng and Karpathy, Andrej and Khosla, Aditya and Bernstein, Michael and others},
  journal=IJCV,
  year={2015},
}

@inproceedings{zhao2023act,
  title={Learning fine-grained bimanual manipulation with low-cost hardware},
  author={Zhao, Tony Z and Kumar, Vikash and Levine, Sergey and Finn, Chelsea},
  booktitle=RSS,
  year={2023}
}

@inproceedings{he2016deep,
  title={Deep residual learning for image recognition},
  author={He, Kaiming and Zhang, Xiangyu and Ren, Shaoqing and Sun, Jian},
  booktitle=CVPR,
  year={2016}
}

@inproceedings{cheng2025omnivtla,
      title={OmniVTLA: Vision-Tactile-Language-Action Model with Semantic-Aligned Tactile Sensing}, 
      author={Zhengxue Cheng and Yiqian Zhang and Wenkang Zhang and Haoyu Li and Keyu Wang and Li Song and Hengdi Zhang},
      year={2026},
      booktitle=ieeeral,
}

@article{litwinkumar2017optimaldegrees,
title = {Optimal Degrees of Synaptic Connectivity},
journal = {Neuron},
year = {2017},
author = {Ashok Litwin-Kumar and Kameron Decker Harris and Richard Axel and Haim Sompolinsky and L.F. Abbott},
}

@article{xu2025unit,
      title={{UniT}: Data Efficient Tactile Representation with Generalization to Unseen Objects}, 
      author={Zhengtong Xu and Raghava Uppuluri and Xinwei Zhang and Cael Fitch and Philip Glen Crandall and Wan Shou and Dongyi Wang and Yu She},
      year={2025},
      journal=ieeeral,
}

@InProceedings{agrawal2025t3,
  title = 	 {Transferable Tactile Transformers for Representation Learning Across Diverse Sensors and Tasks},
  author =       {Zhao, Jialiang and Ma, Yuxiang and Wang, Lirui and Adelson, Edward},
  booktitle = 	CORL,
  year = 	 {2025},
}

@article{khurana2025sarl,
  title={SARL: Spatially-Aware Self-Supervised Representation Learning for Visuo-Tactile Perception},
  author={Khurana, Gurmeher and Wei, Lan and Zhang, Dandan},
  journal={arXiv preprint arXiv:2512.01908},
  year={2025}
}

@InProceedings{higuera2025sparsh,
  title = 	 {Sparsh: Self-supervised touch representations for vision-based tactile sensing},
  author =       {Higuera, Carolina and Sharma, Akash and Bodduluri, Chaithanya Krishna and Fan, Taosha and Lancaster, Patrick and Kalakrishnan, Mrinal and Kaess, Michael and Boots, Byron and Lambeta, Mike and Wu, Tingfan and Mukadam, Mustafa},
  booktitle = 	CORL,
  year = 	 {2025},
}

@inproceedings{anytouch_2026,
 author = {Feng, Ruoxuan and Hu, Jiangyu and Xia, Wenke and Gao, Tianci and Shen, Ao and Sun, Yuhao and Fang, Bin and Hu, Di},
 booktitle = ICLR,
 title = {AnyTouch: Learning Unified Static-Dynamic Representation across Multiple Visuo-tactile Sensors},
 year = {2025}
}

@inproceedings{anytouch2_2026,
 author = {Feng, Ruoxuan and Zhou, Yuxuan and Mei, Siyu and Zhou, Dongzhan and Wang, Pengwei and Cui, Shaowei and Fang, Bin and Yao, Guocai and Hu, Di},
 booktitle = {ICML},
 title = {AnyTouch 2: General Optical Tactile Representation Learning For Dynamic Tactile Perception},
 year = {2026}
}

@INPROCEEDINGS{freetacman2026,
      title={FreeTacMan: Robot-free Visuo-Tactile Data Collection System for Contact-rich Manipulation}, 
      author={Longyan Wu and Checheng Yu and Jieji Ren and Li Chen and Yufei Jiang and Ran Huang and Guoying Gu and Hongyang Li},
      year={2026},
      booktitle=ICRA
}

@INPROCEEDINGS{vital2025,
  author={George, Abraham and Gano, Selam and Katragadda, Pranav and Farimani, Amir Barati},
  booktitle=ICRA, 
  title={VITaL Pretraining: Visuo-Tactile Pretraining for Tactile and Non-Tactile Manipulation Policies}, 
  year={2025},}

@INPROCEEDINGS{convitac2025,
  author={Wu, Zhiyuan and Zhao, Yongqiang and Luo, Shan},
  booktitle=IROS, 
  title={ConViTac: Aligning Visual-Tactile Fusion with Contrastive Representations}, 
  year={2025},}

@INPROCEEDINGS{liu2026crossmodalvisuotactilerepresentationlearning,
      title={Cross-Modal Visuo-Tactile Representation Learning with Action Chunking Transformers for Contact-Rich Manipulation}, 
      author={Yaohua Liu and Rong Fu and Amir H. Gandomi and Simon Fong and Hengjun Zhang},
      year={2026},
      booktitle=RSS, 
}

@ARTICLE{tactilealoha2025,
  author={Gu, Ningquan and Kosuge, Kazuhiro and Hayashibe, Mitsuhiro},
  journal=ieeeral, 
  title={TactileAloha: Learning Bimanual Manipulation With Tactile Sensing}, 
  year={2025},}

@article{huang2026sharpa,
      title={Spatially anchored Tactile Awareness for Robust Dexterous Manipulation}, 
      author={Jialei Huang and Yang Ye and Yuanqing Gong and Xuezhou Zhu and Yang Gao and Kaifeng Zhang},
      year={2026},
      journal={arXiv preprint 2510.14647},
}

@inproceedings{zhao2026vitactracingvisualtactileimitationlearning,
      title={ViTac-Tracing: Visual-Tactile Imitation Learning of Deformable Object Tracing}, 
      author={Yongqiang Zhao and Haining Luo and Yupeng Wang and Emmanouil Spyrakos Papastavridis and Yiannis Demiris and Shan Luo},
      year={2026},
      booktitle=ICRA,
}

@inproceedings{heng2026vitacformer, 
      title={ViTacFormer: Learning Cross-Modal Representation for Visuo-Tactile Dexterous Manipulation}, 
      author={Liang Heng and Haoran Geng and Kaifeng Zhang and Pieter Abbeel and Jitendra Malik},
      year={2026},
      booktitle=RSS,
}

@article{ruan2026retacact,
      title={ReTac-ACT: A State-Gated Vision-Tactile Fusion Transformer for Precision Assembly}, 
      author={Minchi Ruan and LiangQing Zhou and Hongtong Li and Zongtao Wang and ZhaoMing Lu and Jianwei Zhang and Bin Fang},
      year={2026},
  journal={arXiv preprint arXiv:2603.09565},
}

@article{gelfusion2025,
      title={GelFusion: Enhancing Robotic Manipulation under Visual Constraints via Visuotactile Fusion}, 
      author={Shulong Jiang and Shiqi Zhao and Yuxuan Fan and Peng Yin},
      year={2025},
      journal={ arXiv preprint 2505.07455},
      }

@article{zhang2026tacvla,
      title={TacVLA: Contact-Aware Tactile Fusion for Robust Vision-Language-Action Manipulation}, 
      author={Kaidi Zhang and Heng Zhang and Zhengtong Xu and Zhiyuan Zhang and Md Rakibul Islam Prince and Xiang Li and Xiaojing Han and Yuhao Zhou and Arash Ajoudani and Yu She},
      year={2026},
      journal={arXiv preprint 2603.12665},
}

@article{huang2025tactilevla,
    author  = {Jialei Huang and Shuo Wang and Fanqi Lin and Yihang Hu and Chuan Wen and Yang Gao},
    title   = {TACTILE-VLA: UNLOCKING VISION-LANGUAGE-ACTION MODEL'S PHYSICAL KNOWLEDGE FOR TACTILE GENERALIZATION},
    journal = {arXiv preprint 2507.09160},
    year    = {2025}
}

@article{huang2026tafvla,
      title={TaF-VLA: Tactile-Force Alignment in Vision-Language-Action Models for Force-aware Manipulation}, 
      author={Yuzhe Huang and Pei Lin and Wanlin Li and Daohan Li and Jiajun Li and Jiaming Jiang and Chenxi Xiao and Ziyuan Jiao},
      year={2026},
      journal={arXiv preprint 2601.20321}, 
}

@article{luu2026manifeel,
      title={ManiFeel: Benchmarking and Understanding Visuotactile Manipulation Policy Learning}, 
      author={Quan Khanh Luu and Pokuang Zhou and Zhengtong Xu and Zhiyuan Zhang and Qiang Qiu and Yu She},
      year={2026},
      journal={arXiv preprint 2505.18472}, 
}

@article{zorin2026taco,
      title={TacO: Benchmarking Tactile Sensors for Object Manipulation}, 
      author={Anya Zorin and Zilin Si and Myungsun Park and Junsung Park and Alexiy Buynitsky and Sachin Bhadang and Taejun Park and Sohee John Yoon and Yong-Lae Park and Oliver Kroemer and Zeynep Temel and Michael T. Tolley and Sha Yi and Xiaolong Wang},
      year={2026},
      journal={arXiv preprint 2605.21976},
}

@article{saka2026contactcontactawaretactilelearning,
      title={CONTACT: CONtact-aware TACTile Learning for Robotic Disassembly}, 
      author={Yosuke Saka and Jyun-Chi Hu and Adeesh Desai and Zhiyuan Zhang and Bihao Zhang and Quan Khanh Luu and Md Rakibul Islam Prince and Minghui Zheng and Yu She},
      year={2026},
      journal={arXiv preprint 2603.08560},
}

@article{zheng2026omnivtavisuotactileworldmodeling,
      title={OmniVTA: Visuo-Tactile World Modeling for Contact-Rich Robotic Manipulation}, 
      author={Yuhang Zheng and Songen Gu and Yupeng Zheng and Weize Li and Yujie Zang and Shuai Tian and Xiang Li and Ce Hao and Chen Gao and Si Liu and Haoran Li and Yilun Chen and Shuicheng Yan and Wenchao Ding},
      year={2026},
      journal={arXiv preprint 2603.19201},
}

@inproceedings{xue2025reactive,
  title     = {Reactive Diffusion Policy: Slow-Fast Visual-Tactile Policy Learning for Contact-Rich Manipulation},
  author    = {Xue, Han and Ren, Jieji and Chen, Wendi and Zhang, Gu and Fang, Yuan and Gu, Guoying and Xu, Huazhe and Lu, Cewu},
  booktitle = RSS,
  year      = {2025}
}

@article{chen2026implicitrdp,
  title     = {Implicitrdp: An end-to-end visual-force diffusion policy with structural slow-fast learning},
  author    = {Chen, Wendi and Xue, Han and Wang, Yi and Zhou, Fangyuan and Lv, Jun and Jin, Yang and Tang, Shirun and Wen, Chuan and Lu, Cewu},
  journal   = ieeeral,
  year      = {2026},
}

@ARTICLE{foar2025,
  author={He, Zihao and Fang, Hongjie and Chen, Jingjing and Fang, Hao-Shu and Lu, Cewu},
  journal=ieeeral, 
  title={FoAR: Force-Aware Reactive Policy for Contact-Rich Robotic Manipulation}, 
  year={2025},}

@inproceedings{alain2017probes,
  title     = {Understanding Intermediate Layers Using Linear Classifier Probes},
  author    = {Alain, Guillaume and Bengio, Yoshua},
  booktitle = {International Conference on Learning Representations (ICLR), Workshop Track},
  year      = {2017},
  note      = {arXiv:1610.01644}
}

@ARTICLE{tacsl,
  author={Akinola, Iretiayo and Xu, Jie and Carius, Jan and Fox, Dieter and Narang, Yashraj},
  journal=ieeetro,
  title={TacSL: A Library for Visuotactile Sensor Simulation and Learning},
  year={2025},}

@ARTICLE{digit,
  author={Lambeta, Mike and Chou, Po-Wei and Tian, Stephen and Yang, Brian and Maloon, Benjamin and Most, Victoria Rose and Stroud, Dave and Santos, Raymond and Byagowi, Ahmad and Kammerer, Gregg and Jayaraman, Dinesh and Calandra, Roberto},
  journal=ieeeral, 
  title={{DIGIT}: A Novel Design for a Low-Cost Compact High-Resolution Tactile Sensor With Application to In-Hand Manipulation}, 
  year={2020},
}

@Article{gelsight,
AUTHOR = {Yuan, Wenzhen and Dong, Siyuan and Adelson, Edward H.},
TITLE = {{GelSight}: High-Resolution Robot Tactile Sensors for Estimating Geometry and Force},
JOURNAL = {Sensors},
YEAR = {2017},
}

\end{document}